\documentclass[11pt,a4paper]{article}

\usepackage[]{ACL2023}

\usepackage{nert}

\usepackage{times}
\usepackage{latexsym}

\usepackage{url}
\usepackage[utf8]{inputenc}
\usepackage{verbatim}
\usepackage{amsmath}

\usepackage{colortbl}
\usepackage{multirow}
\usepackage{caption}
\usepackage{booktabs}

\usepackage{graphicx}
\usepackage{pgf}

\usepackage{rotating}

\usetikzlibrary{bayesnet}

\usepackage{relsize}

\title{Reanalyzing L2 Preposition Learning with Bayesian Mixed Effects \\and a Pretrained Language Model}

\author{Jakob Prange \\
  Hong Kong Polytechnic University \\
  {\tt jakob.prange@polyu.edu.hk} \\\And
  Man Ho Ivy Wong \\
  Hong Kong Shue Yan University \\
   {\tt mhwong@hksyu.edu} \\
  }

\begin{document}

\maketitle

\begin{abstract}
We use both Bayesian and neural models to dissect a data set of Chinese learners' pre- and post-interventional responses to two tests measuring their understanding of English prepositions.
The results mostly replicate previous findings from frequentist analyses and reveal new and crucial interactions between student ability, task type, and stimulus sentence.
Given the sparsity of the data as well as high diversity among learners, the Bayesian method proves most useful; but we also see potential in using language model probabilities as predictors of grammaticality and learnability.%
\footnote{Our experimental code is available at \url{https://github.com/jakpra/L2-Prepositions}.}
\end{abstract}

\section{Introduction}

Learning a second or third language is hard---not only for NLP models but also for humans!
Which linguistic properties and external factors make it so difficult? And how can we improve instruction and testing to help learners accomplish their goals?
Here we also ask a third question: How can we best apply different computational models to such behavioral experimental data in order to get intuitive and detailed answers to the first two questions in a practical and efficient way?
For example, we are interested in whether language model (LM) probabilities might give a rough estimate of grammaticality and learning difficulty (\cref{tab:examples-probs}, right columns).

This work is in part a replication study of \citet{wong2022fostering}, who, in addressing these questions about native Chinese speakers' learning of English prepositions in context (see examples in \cref{tab:examples-probs}), mainly focused on instructional intervention and found generally positive effects as well as differences between instruction types, in particular favoring conceptual over rule-based teaching.
We pick up where \citet{wong2022fostering} left off and search for more fine-grained patterns among students' individual differences, linguistic items, stimulus sentences, and their grammaticality.
Our main hypothesis is that the full story of the complex interactions among these factors can only be revealed by modeling them holistically.
%
Such a fine-grained holistic analysis is well-aligned with Item Response Theory \citep[IRT;][]{FISCHER1973359,lord1980applications}.
IRT allows us to formulate models in terms of predicting whether students provide the intended response to each test item. 
We consider sparse Bayesian and dense neural versions of this framework.
%
We can then inspect how strongly each model input (the linguistic, experimental, and student-specific factors mentioned above, which are realized as random and fixed effects for the Bayesian model and feature vectors for the neural model) affects the outcome.
As a representative of yet another modeling strategy, and also as an additional input to the IRT models, we obtain probability estimates for the target prepositions in each stimulus sentence from a pretrained transformer LM.
These probabilities serve as a proxy for contextual formulaicity, as learned distributionally from a large corpus.

%
%

\begin{table*}[ht]
    \setlength{\tabcolsep}{2.9pt}
    \centering\small
    \begin{tabular}{c c lccrrcc}
    &&&&\multicolumn{2}{c}{\textbf{Student}}&& \\
    &&&&\multicolumn{2}{c}{\textbf{judgment}} & \multicolumn{2}{c}{\textbf{LM}} & \\\cmidrule(lr){5-6}\cmidrule(l){7-8}
\# & Usage & \multicolumn{2}{l}{Stimuli \hspace*{6.3cm} Grammatical?} & pre & \multicolumn{1}{c}{post} & \multicolumn{1}{c}{$p_{tgt}$} & $p_{ctx}$ & \\\cmidrule(r){1-4}\cmidrule(lr){5-6}\cmidrule(l){7-8}
1a & \textsc{Hir}-Spat & The bell hung \textbf{over}\phantom{ugh} the baby's cradle and made him smile.  & $\checkmark$ & 80.65 & 92.59 & \cellcolor{yellow!50} \textbf{4.15} & \cellcolor{yellow!50} \textbf{54.66} & \parbox[t]{2mm}{\multirow{6}{*}{\rotatebox[origin=c]{90}{As expected}}} \\
1b & & \phantom{The bell hung }\textbf{through}\phantom{ the baby's cradle and made him smile.}  & \ding{55} & 50.00 & 46.67 &  0.03 & 54.10 & \\\cmidrule(r){1-4}\cmidrule(lr){5-6}\cmidrule(l){7-8}
2a & \textsc{Hir}-Abst & The tutors watched \textbf{over} the students during the oral presentation. & $\checkmark$ & 80.00 & 96.77 &  \cellcolor{yellow!50} \textbf{96.70} & \cellcolor{yellow!50} \textbf{67.19} & \\
2b & & \phantom{The tutors watched }\textbf{on}\phantom{ver the students during the oral presentation.} &  \ding{55} & 35.00 & 46.15 & 0.07 & 50.64 & \\\cmidrule(r){1-4}\cmidrule(lr){5-6}\cmidrule(l){7-8}
3a & \textsc{Cvr}-Abst & Tremendous fear fell \textbf{over}\phantom{ugh} the town after the murder.  & $\checkmark$ & 71.43 & 91.67 &  \cellcolor{yellow!50} \textbf{18.58} & \cellcolor{yellow!50} \textbf{65.61} & \\
3b & & \phantom{Tremendous fear fell }\textbf{through}\phantom{ the town after the murder.}  & \ding{55} & 41.67 & 27.27 &  0.39 & 63.91 & \\
\midrule\midrule
4a & \textsc{Crs}-Spat & The painter reached \textbf{over}\phantom{ugh} the paint can for a brush.  & $\checkmark$ & 41.18 & 63.33 & \cellcolor{red!35} {0.05} & \cellcolor{yellow!50} \textbf{49.50} & \parbox[t]{2mm}{\multirow{6}{*}{\rotatebox[origin=c]{90}{$\operatorname{sign}\left(\Delta p_{tgt}\right)$ ???}}} \\
4b & & \phantom{The painter reached }\textbf{through}\phantom{ the paint can for a brush.}  & \ding{55} & 36.11 & 33.33 &  \cellcolor{red!35} \textbf{0.78} & 45.42 & \\\cmidrule(r){1-4}\cmidrule(lr){5-6}\cmidrule(l){7-8}
5a & \textsc{Crs}-Abst & The lawyer jumped \textbf{over} a few pages of the contract.  & $\checkmark$ & 72.73 & 94.12 & \cellcolor{red!35} 6.97 & \cellcolor{yellow!50} \textbf{52.39} & \\
5b & & \phantom{The lawyer jumped }\textbf{to}\phantom{ver a few pages of the contract.}  & \ding{55} & 42.11 & 34.78 &  \cellcolor{red!35} \textbf{20.27} & 51.66 & \\\cmidrule(r){1-4}\cmidrule(lr){5-6}\cmidrule(l){7-8}
6a & \textsc{Cvr}-Abst & Happiness diffused \textbf{over} the guests when they see the newly-weds.  & $\checkmark$ & 44.44 & 88.46 & \cellcolor{red!35} {2.47} & \cellcolor{yellow!50} \textbf{65.27} & \\
6b & & \phantom{Happiness diffused }\textbf{on}\phantom{ver the guests when they see the newly-weds.}  & \ding{55} & 80.00 & 35.48  & \cellcolor{red!35} \textbf{3.19} & 62.82 & \\
\midrule\midrule
7a & \textsc{Cvr}-Spat & The canvas stretched \textbf{over}\phantom{ugh} a large hole in the road.  & $\checkmark$ & 44.12 & 70.37 &  \cellcolor{yellow!50} \textbf{17.99} & \cellcolor{red!35} {51.66} & \parbox[t]{2mm}{\multirow{6}{*}{\rotatebox[origin=c]{90}{$\operatorname{sign}\left(\Delta p_{ctx}\right)$ ???}}} \\
7b & & \phantom{The canvas stretched }\textbf{through}\phantom{ a large hole in the road.}  & \ding{55} & 55.56 & 60.00 &  15.52 & \cellcolor{red!35} \textbf{52.79} & \\\cmidrule(r){1-4}\cmidrule(lr){5-6}\cmidrule(l){7-8}
8a & \textsc{Cvr}-Abst & The tension swept \textbf{over} the school when the alarm rang.  & $\checkmark$ & 66.67 & 100.00 &  \cellcolor{yellow!50} \textbf{3.93} & \cellcolor{red!35} 46.61 & \\
8b & & \phantom{The tension swept }\textbf{onto}\phantom{ the school when the alarm rang.}  & \ding{55} & 37.84 & 12.50 &  <0.01 & \cellcolor{red!35} \textbf{46.63} & \\\cmidrule(r){1-4}\cmidrule(lr){5-6}\cmidrule(l){7-8}
9a & \textsc{Crs}-Abst & The politicians skipped \textbf{over} sensitive topics during the debate.  & $\checkmark$ & 83.33 & 94.59 &  \cellcolor{yellow!50} \textbf{2.88} & \cellcolor{red!35} {40.80} & \\
9b & & \phantom{The politicians skipped }\textbf{to}\phantom{ver sensitive topics during the debate.}  & \ding{55} & 60.98 & 35.00 & 0.17 & \cellcolor{red!35} \textbf{42.06} & \\
    \bottomrule
    \end{tabular}
    \caption{Examples of stimulus sentences for grammatical ($\checkmark$) and ungrammatical (\ding{55}) preposition use. In the \textbf{Student judgment} columns we show the percentage of students who judged the example as grammatical at the pretest (including control group) and posttest (treatment groups only) in \citeauthor{wong2022fostering}'s study. 
    The \textbf{LM} columns show our probed RoBERTa probabilities $p_{tgt}$ and $p_{ctx}$ [in \%], which are defined in \cref{sec:roberta} and discussed in \cref{sec:exp-lm-prob}.
    }
    \label{tab:examples-probs}
\end{table*}

While the theoretical advantages of Bayesian over frequentist statistics, as well as the generally strong performance of neuro-distributional models, are often cited as justification for choosing one particular modeling method, both replication studies and side-by-side comparisons of such drastically different modeling strategies for linguistic analysis remain rare \citep[with notable exceptions, e.g.,][]{michaelov2023can,tack2021mark}.
%
%
%

We contribute to this important development by
\begin{itemize}
    \item designing (\cref{sec:blm}), fitting, and evaluating (\cref{sec:replication}) a Bayesian mixed effects model on \citeposs{wong2022fostering} data (\cref{sec:data}), considering more potential linguistic and human factors in preposition learning than previously and finding significant effects for several of them;
    \item training (\cref{sec:mlp}) and evaluating an analogous multilayer perceptron (MLP) model and comparing it with the Bayesian model in terms of both feature ablation and overall prediction accuracy of the outcome, i.e., whether a student will answer a test prompt correctly (\cref{sec:prediction});
    \item and probing a pretrained LM (\cref{sec:roberta,sec:exp}) for contextual probabilities of target prepositions in order to determine their correlation---and thus, practical usefulness---with human language learning.
\end{itemize}

Thus, we aim to both better explain L2 preposition learning and compare Bayesian, frequentist, and neural approaches to doing so.






 



\section{Background}\label{sec:background}

\subsection{English Preposition Semantics}

Prepositions are among the most frequently used word classes in the English language---they make up between 6 and 10 \% of all word tokens depending on text type and other factors \citep[cf.][]{schneider2018comprehensive}.
This is because English does not have a full-fledged morphological case system and instead often expresses semantic roles via word order and lexical markers like prepositions.
At the same time, the inventory of preposition forms is relatively small---a closed set of largely grammaticalized function words covering a wide range of predictive, configurational, and other relational meanings.
The resulting many-to-many mapping between word forms and meanings is complex and
%
%
warrants nuanced linguistic annotation, analysis, and computational modeling in context \citep{ohara-03,hovy-10,srikumar-13,schneider2018comprehensive,kim2019predicting}.
Further, considerable cross-linguistic variation in the precise syntax-semantics interactions of prepositions and case has been shown to affect not only machine translation \citep{hashemi2014comparison,weller-etal-2014-using,popovic2017comparing}, but also construal in human translation \citep{hwang-etal-2020-k,peng_corpus_2020,prange-schneider-2021-german} and---crucially---learner writing \citep{littlemore2006figurative,mueller-11,gvarishvili2013interference,kranzlein-etal-2020-pastrie}.

\subsection{Cognitive and Concept-based Instruction}

Cognitive linguistics (CogLx) maintains that many aspects of natural language semantics are grounded in extra-linguistic cognition, even (or especially) when they do not directly arise from syntactic composition, or at the lexical level.
For example, \citet{brugman-88}, \citet{lakoff-87}, and \citet{tyler-03} argue that spatial prepositions evoke a network of interrelated senses, ranging from more prototypical to extended and abstract ones. Incorporating such conceptual connectedness into language instruction has shown some benefits \citep{tyler-12,boers-98,lam-09}.

\subsection{Computational Modeling in SLA}\label{sec:model-backgr}

Until recently, most studies in applied linguistics and second-language acquisition (SLA)---insofar as they are quantitative---have relied on null-hypothesis testing with frequentist statistical measurements like analysis of variance (ANOVA) \citep{norouzian2018bayesian}.
This has the advantage that it is generally unambiguous and interpretable what is being tested (because concrete and specific hypotheses need to be formulated ahead of time) and that conclusions are based directly on data without any potentially confounding modeling mechanisms. 
At the same time, frequentist analyses are relatively rigid, and thus run into efficiency, sparsity, and reliability issues as interactions of interest grow more complex.
\citet{li2022digital} propound a more widespread use of computational modeling and AI in language learning and education research.
A promising alternative exists in the form of Bayesian models \citep[e.g.,][]{murakami2022effects,privitera_momenian_weekes_2022,guo-21,norouzian2018bayesian,norouzian2019bayesian}, which circumvent sparsity by sampling from latent distributions and offer intuitive measures of uncertainty ``for free'' in form of the estimated distributions' scale parameters.
They can also be made very efficient to train by utilizing stochastic variational inference (SVI).

Bayesian modeling for educational applications goes hand-in-hand with Item Response Theory \citep[IRT;][]{FISCHER1973359,lord1980applications}, which posits that learning outcomes depend on both student aptitude and test item difficulty. This addresses another limitation of frequentist analysis---the focus on aggregate test scores---by modeling each student's response to each item individually.
We loosely follow this general paradigm with our model implementations, without committing to any specific theoretical assumptions.

Within NLP, Bayesian and IRT-based approaches have been used to evaluate both human annotators \citep{rehbein-ruppenhofer-2017-detecting,passonneau-14} and models \citep{kwako-etal-2022-using,sedoc-ungar-2020-item}, to conduct text analysis \citep{kornilova-etal-2022-item,bamman-etal-2014-bayesian,wang-etal-2012-historical}, and natural language inference \citep{gantt-etal-2020-natural}.

\citet{murakami2022effects} show that grammar learning can be affected by contextual predictability (or formulaicity).
While they used a simple $n$-gram model, we account for this phenomenon more broadly with a pretrained transformer LM.


\section{Original Study and Data}\label{sec:orig-study}\label{sec:data}

\citet{wong2022fostering} measured students' pre- and post-interventional understanding of the English prepositions \textit{in}, \textit{at}, and \textit{over}, particularly contrasting CogLx\slash schematics-based instruction with different flavors of rule-based methods.
To this end, intermediate learners of English (all university students) with first languages Mandarin or Cantonese took initial English language tests (`pretest') targeting different usages of prepositions. They were then taught with one of four methods (incl. one control group, who received instruction about definite and indefinite articles instead of prepositions), and subsequently tested two more times.
There were two different tests: a grammaticality judgment test (GJT) to measure effects on language processing and a picture elicitation test (PET) to measure effects on production.

While all preposition-focused training was found to enhance learners' understanding of prepositions compared to both the pretest and the control group, schematics-based mediation led to stronger learning results
than any of the other methods, 
especially at the PET (\cref{fig:effect-results-repl}) and on spatial usages of prepositions (the interaction between instruction method and spatial usage is not shown in \cref{fig:effect-results-repl} for brevity).
These latter findings in particular support our hypothesis that in addition to external factors like task type and instruction method, learning difficulty may also be affected by \textit{inherent linguistic} properties of the prepositions and their usages (just as, e.g., \citet{guo-21} show for distributional properties of grammatical suffixes).
In this work we take a second look at \citeauthor{wong2022fostering}'s data to directly address this possibility for preposition learning.



\subsection{Data Summary}

We conduct all of our computational analyses with \citeauthor{wong2022fostering}'s data (stimuli and behavioral results)
but expand on the original study by explicitly modeling as potential factors several additional dimensions, relating to individual differences and interactions among stimuli, task types, and students (\cref{tab:factors}, \cref{sec:stimuli,sec:mod-factors}).
71 students (after outlier filtering) participated in the study.
There are a total of 48 test items (12 senses $\times$ 4 contexts) and 22 fillers for the GJT as well as 36 test items (12 senses $\times$ 3 contexts) and 15 fillers for the PET.
Outlier students and filler items are removed before any analysis\slash model training, resulting in 17,644 data points overall (GJT: 10,156; PET: 7,488).

\subsection{Stimulus Sentences}\label{sec:stimuli}
In the GJT (but not in the PET), students receive a linguistic stimulus to evaluate for grammaticality (see examples in \cref{tab:examples-probs}). Intended-grammatical stimuli involve target prepositions used in a sentence context that evokes their intended sense or function (fxn), either literally\slash spatially or figuratively\slash abstractly.
For each intended-grammatical stimulus, there is an intended-\textbf{\textit{un}}grammatical stimulus, consisting of the same sentence context but replacing the target preposition with another that is meant to fit the context less well.

\subsection{Categorical Features}\label{sec:mod-factors}

\noindent\textbf{Instruction method.}
The main goal of \citeposs{wong2022fostering} study was to compare CogLx-based schematic mediation (\textit{SM}) with more traditional rule-and-exemplar (\textit{RM}) and bare-bones correctness-based mediation (\textit{CM}).  
SM, RM, and CM instruction focused on the same preposition forms and usages students were tested on.

\noindent\textbf{Time of test.}
Students were tested three times: Two days before instructional intervention (\textit{PRE}test, $\triangleleft$ in \cref{fig:effect-results-repl}), two days after instruction (\textit{POST}test, $\circ$), and again 3 weeks later (\textit{D}e\textit{L}a\textit{Y}ed posttest, $\triangleright$).

\noindent\textbf{Preposition form, function (fxn), and usage.}
The test cues are made up of 6 pairs of preposition usages across three forms: `\textit{in}' with the \textsc{Containment} (\textsc{Ctn}) function; `\textit{at}' with the \textsc{Target} (\textsc{Tgt}) and \textsc{Point} (\textsc{Pnt}) functions; and `\textit{over}' with the \textsc{Higher} (\textsc{Hir}), \textsc{Across} (\textsc{Crs}), and \textsc{Cover} (\textsc{Cvr}) functions. 
Each usage pair consists of a spatial (e.g., `in the box') and a non-spatial cue (e.g., `in love') sharing the same schematization (in this case, \textsc{Containment}).
The cues were selected based on the Principled Polysemy Framework \citep{tyler-03}, thereby ruling out overly fine-grained senses and allowing systematic presentation for instruction and testing.

\noindent\textbf{Test type.}
In the \textit{GJT}, learners had to decide, for each stimulus sentence containing a preposition, whether the whole sentence is ``correct'' or ``incorrect''.\footnotemark\
We consider as a potential factor on the outcome whether a stimulus is intended-grammatical (\textit{GJT-Y}) or not (\textit{GJT-N}).
In the \textit{PET}, learners were shown an illustration of a concrete scenario instantiating one of the cues and were asked to produce a descriptive sentence containing a preposition. Responses were counted as correct if they chose the target preposition.

\footnotetext{The testing prompt did not explicitly highlight or otherwise draw attention to the preposition in question.}

\noindent\textbf{Students.}
By adding local student identities to the model input (anonymized as, e.g., $s_1$, $s_{23}$), we allow fine-grained degrees of freedom w.r.t.\ individual differences, as is suggested by IRT.

\begin{table}[t]
    \setlength{\tabcolsep}{2.5pt}
    \centering\small
    \begin{tabular}{l l cc}
    & & W22 & Ours \\\midrule
        \multicolumn{2}{l}{\textbf{Random Effects}} \\  
        \textbf{Feature} & \textbf{Values} \\  
        Instruction & \textit{SM, RM, CM, CTRL} & $\checkmark$ & $\checkmark$ \\
        Time & \textit{PRE, POST, DLY} & $\checkmark$ & $\checkmark$ \\
        Test & \textit{GJT, PET} & $\checkmark$ & $\checkmark$ \\
        Usage & \textit{Spatial, Abstract} & $\checkmark$ & $\checkmark$ \\
        Answer & \textit{GJT-Y, GJT-N, PET} & \ding{55} & $\checkmark$ \\
        Form-Fxn & \textit{in}-\textsc{Ctn}, \textit{at}-\textsc{Tgt} & \ding{55} & $\checkmark$ \\
                               & \textit{at}-\textsc{Pnt}, \textit{over}-\textsc{Hir}, \\
                               & \textit{over}-\textsc{Crs}, \textit{over}-\textsc{Cvr} \\
        Student & $s_1$, ..., $s_{71}$ & \ding{55} & $\checkmark$ \\\midrule
        \multicolumn{2}{l}{\textbf{Fixed Effects}} \\  
        \multicolumn{2}{l}{$p_{tgt}$---LM probability of target preposition} & \ding{55} & $\checkmark$ \\
        \multicolumn{2}{l}{$p_{ctx}$---Avg. LM prob. of non-tgt tokens in sent.} & \ding{55} & $\checkmark$ \\
    \bottomrule
    \end{tabular}
    \caption{Features under consideration in \citet{wong2022fostering} (W22) and our work.}
    \label{tab:factors}
\end{table}

\section{Models}\label{sec:models}

Our main point of reference (or quasi-baseline) is \citeauthor{wong2022fostering}'s frequentist data analysis, which is summarized in \cref{sec:orig-study}. 
In this work, we newly consider the following different modeling strategies:
We train a \textbf{Bayesian logistic model} (\textbf{BLM}, \cref{sec:blm}) as well as a small \textbf{multilayer perceptron} (\textbf{MLP}, \cref{sec:mlp}) on the same data.
With the BLM we can define and interpret the precise structure of how individual features and their interactions affect the outcome. In contrast, the MLP utilizes nonlinear activation functions and multiple iterations\slash layers of computation, allowing it to pick up on complex interactions among input features without prior specification and thus to potentially achieve higher predictive accuracy, at the cost of interpretability.
Both the BLM and MLP are implemented in Python and PyTorch, and are light-weight enough to be trained and run on a laptop CPU within several minutes for training and several seconds for inference.
We also query a pretrained \textbf{neural language model} \citep[\textbf{LM}, namely RoBERTa;][]{liu-19-1} to obtain contextual probabilities for the stimulus sentences used in the grammaticality judgment test and add those probabilities to the BLM and MLP's inputs (\cref{sec:roberta}).

\subsection{Bayesian Logistic Model}\label{sec:blm}

We model the posterior likelihood of a correct response (i.e., a given student providing the intended answer to a given stimulus) as a logistic regression conditional on the aforementioned categorical variables.
Concretely, responses are sampled from a Bernoulli distribution with log-odds proportional to the weighted sum of the random and fixed effects.
As potential factors we consider the features listed in \cref{sec:mod-factors,tab:factors}, as well as their mutual interactions. For the \textit{students} feature, to keep model size manageable, we only consider pairwise interactions with usage (spatial\slash abstract), form-fxn, and answer. Otherwise all $n$-wise interactions are included.
The effects' weight coefficients are sampled from Normal distributions whose means and standard deviations are fitted to the training data via SVI with the AdamW optimizer, AutoNormal guide, and ELBO loss.  
We use standard-normal priors for means and flat half-normal priors for standard deviations, meaning that, by default, parameter estimates are pulled towards null-effects, and will only get more extreme if there is strong evidence for it.
The model is implemented using the Pyro-PPL\slash BRMP libraries \citep{bingham2018pyro}. 

\subsection{Multilayer Perceptron}\label{sec:mlp}

We train and test a multilayer perceptron (MLP) with depth 3.
We mirror the BLM setup by treating student response correctness as the output and optimization objective and the different feature sets as concatenated embedding vectors.
Between hidden layers we apply the GELU activation function, and during training additionally dropout with $p=0.2$ before activation. We also apply dropout with $p=0.1$ to the input layer.
We minimize binary cross-entropy loss using the AdamW optimizer.
We train for up to 25 epochs but stop early if dev set accuracy does not increase for 3 consecutive epochs.

\subsection{RoBERTa}\label{sec:roberta}

We feed the GJT stimulus sentences to RoBERTa-base \citep[][accessed via Huggingface-transformers]{liu-19-1}. RoBERTa a pretrained neural LM based on the transformer architecture \citep{vaswani-17} and trained on English literary and Wikipedia texts to optimize the masked-token and next-sentence prediction objectives.
For each sentence, we wish to obtain RoBERTa's posterior probability estimates for each observed word token $w_i \in \textbf{w}_{0:n-1}$, given $\textbf{w}_{0:n-1} \textbackslash \{w_i\}$, i.e., all other words in that sentence. Thus we run RoBERTa $n$ times, each time $i$ masking out $w_i$ in the input.
From these $n$ sets of probabilities, we extract two measurements of formulaicity we expect to be relevant to our modeling objective of student response correctness:\footnote{We also preliminarily experimented with inputting the entire LM hidden state of the last layer to the models but did not find it to be helpful. \citet{kauf-2022} found that alignment with human judgments varies from layer to layer, which presents an interesting avenue for future work.}
\begin{inparaenum}[(a)]
\item $p_{tgt}$, the contextual probability of the target or alternate preposition given all other words in the sentence and 
\item $p_{ctx}$, the average contextual probability of all words \textit{except} the preposition.\footnote{Note that the preposition token still has the potential to affect the other words' probabilities by occurring in their context condition.}
\end{inparaenum}
Examples are given in \cref{tab:examples-probs}.
We standardize these two variables to $\mathcal{N}(0, 1)$ and add them to the BLM (as fixed effects, both individually and with interactions) and MLP (as scalar input features).

\section{Evaluation}\label{sec:exp}

We first analyze the BLM's learned latent coefficients (\cref{sec:exp-interaction}).
Then we compare different versions of the BLM and MLP w.r.t.\ their ability to predict unseen student responses using their estimated weighting of linguistic and external features as well as LM probabilities (\cref{sec:prediction}).
Finally, we manually inspect a small set of stimulus sentences with anomalous LM probabilities w.r.t.\ their intended grammaticality and observed judgments (\cref{sec:exp-lm-prob}).

\subsection{Determining Relevant Input Features}\label{sec:replication}\label{sec:exp-interaction}

\noindent\textbf{Setup.}
We fit BLMs on the entire data set (without reserving dev or eval splits).
We run SVI for 1000 iterations with a sample size of 100 and a fixed random seed.  
We compute effect sizes (Cohen's $d$), and $p$-values based on 95\%-confidence intervals of differences between estimated parameter values \citep{altman2011obtain}.



\noindent\textbf{Replication.}
As in \citet{wong2022fostering}, we use the features \textit{instruction, time, form-fxn, usage}, and additionally let the model learn individual coefficients for each student. Separate models were trained for GJT and PET.
As shown in \cref{fig:effect-results-repl}, we mostly replicate similar trends (differences between differences) as found previously, namely:


\begin{itemize}
    \item \textit{Time:} DLY $\approx$ POST $>$ PRE;
    \item \textit{Instruction:} treatment $>$ ctrl; SM $>$ CM $\approx$ RM;
    \item and we generally see larger learning effects in the PET than in the GJT.
\end{itemize}

However, many effect sizes are amplified---and thus $p$-values more significant-looking---in our model.
A potential explanation for this could be that the BLM models each individual item response whereas ANOVA only considers overall \%-correct. We are thus comparing effects on all students' accuracy at multiple test items in aggregate with effects on each student's accuracy at each test item separately. It seems intuitive that the latter `micro-effects' are much greater on average than the former `macro-effects', which are themselves effects on the average performance metric.
Another reason could be that because the Bayesian effect sizes stem from simulated data points, they are only indirectly related to the real underlying data via SVI. The estimated distribution these samples are drawn from only approximates the real data and thus the effect size estimations may be over-confident. See \cref{sec:model-discussion} for a discussion of advantages and disadvantages.

Although our model estimates spatial usages as generally more difficult than abstract ones, we do not replicate \citeauthor{wong2022fostering}'s finding of an \textit{interaction} between abstractness and instruction or time.
Still, our Bayesian quasi-IRT approach allows us to find additional interesting patterns that could not have been captured by a frequentist analysis\footnote{Or only very tediously so.} as they involve student-level and item-level interactions:

\begin{figure}[p]
    \centering\small
    \includegraphics[width=0.605\columnwidth]{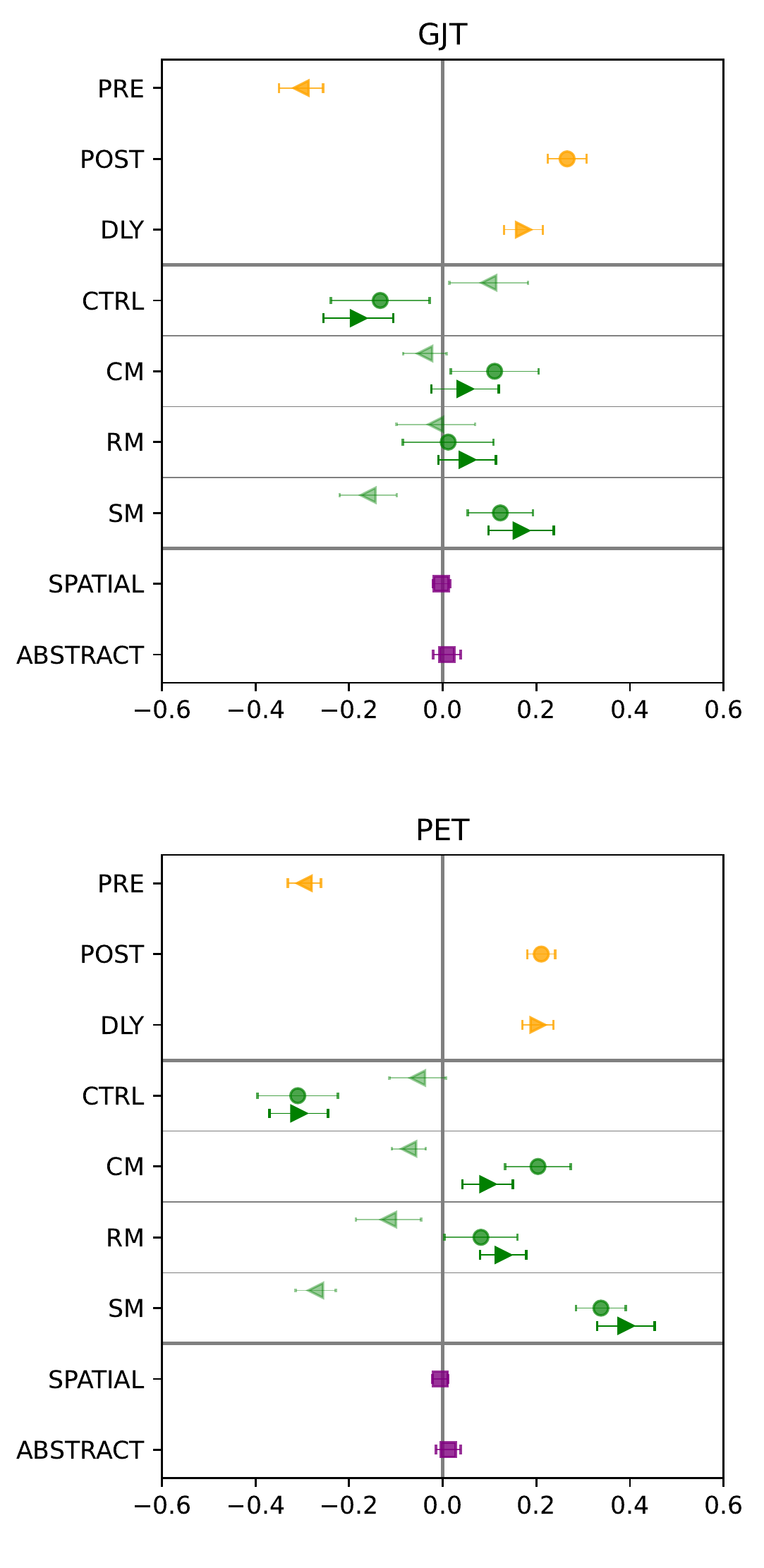}\hspace*{-0.02\columnwidth}%
    \includegraphics[width=0.395\columnwidth]{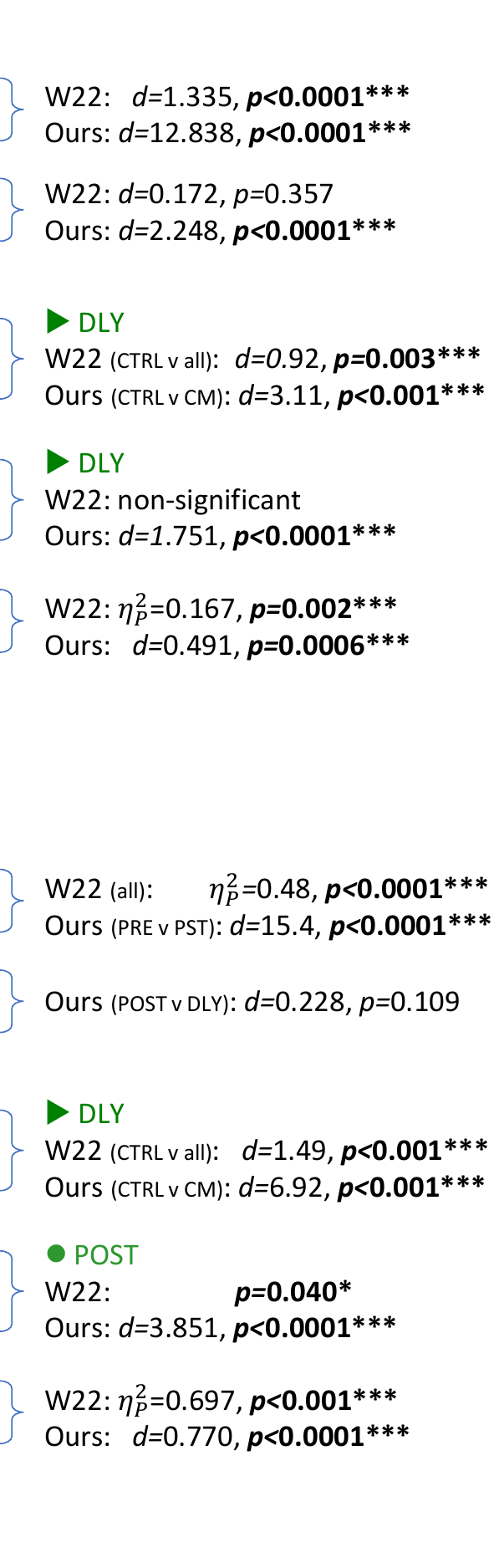}
    \caption{Summary of our Bayesian effect estimations (marginal means and standard deviations over model parameters) for selected features. Coefficient values ($x$-axis) indicate the extent to which the feature value ($y$-axis) contributes to a correct (positive) or incorrect (negative) student response. On the right we compare effect sizes (Cohen's $d$) and statistical significance to \citeposs{wong2022fostering} frequentist analysis.\footnotemark}
    \label{fig:effect-results-repl}
\end{figure}

\begin{figure}[p]
    \includegraphics[width=0.7\columnwidth]{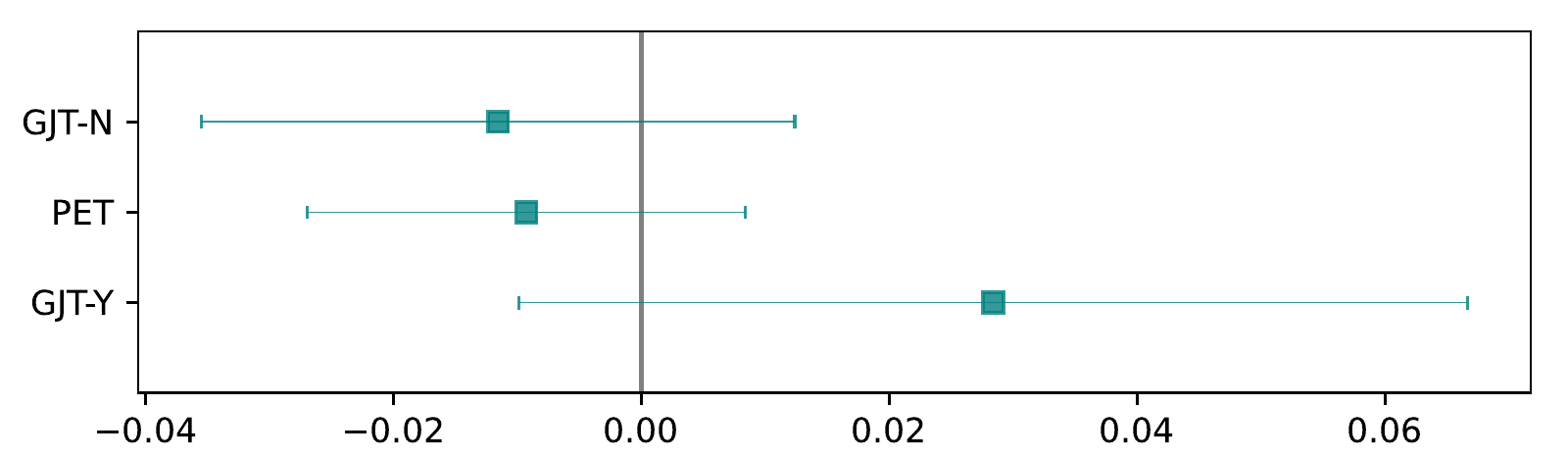}%
    \includegraphics[width=0.29\columnwidth]{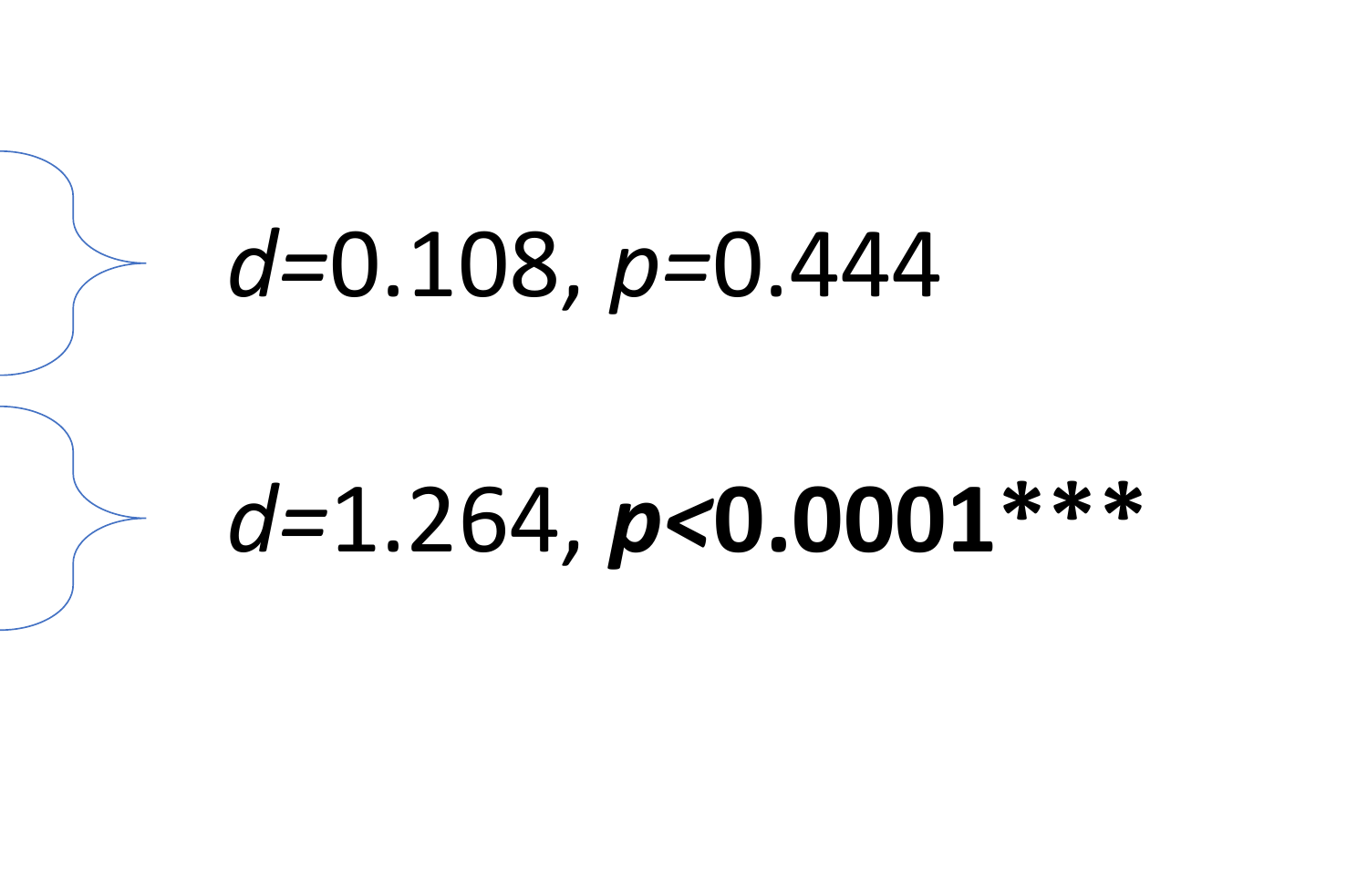}
    \caption{Estimated effects for different answer types.}
    \label{fig:effect-results-answer}
\end{figure}

\begin{figure}[p]
    \includegraphics[width=\columnwidth]{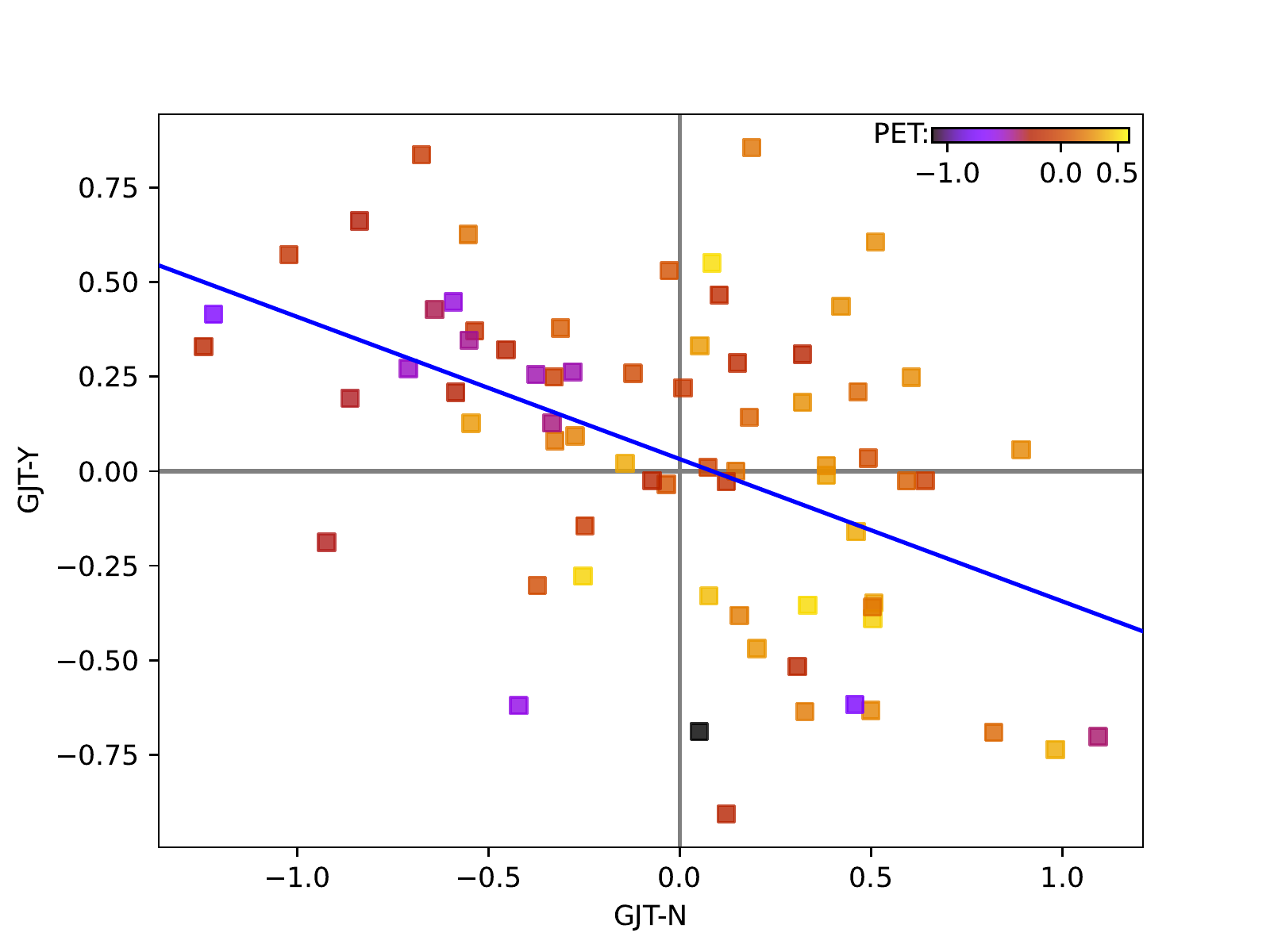}
    \caption{Effect estimation means of individual students (points) in interaction with answer type ($x$=GJT-N, $y$=GJT-Y, color=PET). There is a weak negative correlation between being good at GJT-N and GJT-Y answers (blue line, $R^2$=0.23) and a weak positive correlation between GJT-N and PET skills ($R^2$=0.20).
    }
    \label{fig:effect-results-stud}
\end{figure}

\footnotetext{Where W22 does not report Cohen's $d$, we show their reported partial-eta-squared $\eta_p^2$ instead.}


\noindent\textbf{Answer type and individual differences.}
We trained a single combined model on both GJT and PET data.
As can be expected, in addition to the overall trends (\cref{fig:effect-results-repl}), we also find a strong effect for expected answer type (\cref{fig:effect-results-answer}): the \textit{receptive} task of accepting grammatical items (GJT\mbox{-}Y) is much easier than the \textit{productive} task of choosing the right preposition when describing a picture (PET).
Interestingly, ruling out ungrammatical items (GJT\mbox{-}N) is equally as difficult as the PET.
In addition, outcomes are affected by individual differences between students, and student aptitude heavily depends on answer type (\cref{fig:effect-results-stud}) as well as on preposition form\slash function (\cref{fig:stud-prep} in \cref{app:detailed-results}).
There is some (negative) correlation between individual aptitudes at GJT\mbox{-}N and GJT\mbox{-}Y and some (positive) correlation between GJT\mbox{-}N and PET.
Still, both correlations are weak ($R^2=$ 0.23 and 0.20).

In sum, not only do receptive vs.\ productive task types vary in their overall difficulty (\cref{fig:effect-results-answer}), but the wide spread in individual student differences (\cref{fig:effect-results-stud}) suggests that the skill sets required (let's call them ``sensitivity'' and ``specificity'') are somewhat complementary to each other and tend to be distributed unevenly among students. Each student has a unique combination of them. We discuss this further in \cref{sec:sla-discussion}.




\begin{figure*}[t]
    \includegraphics[width=0.33\textwidth]{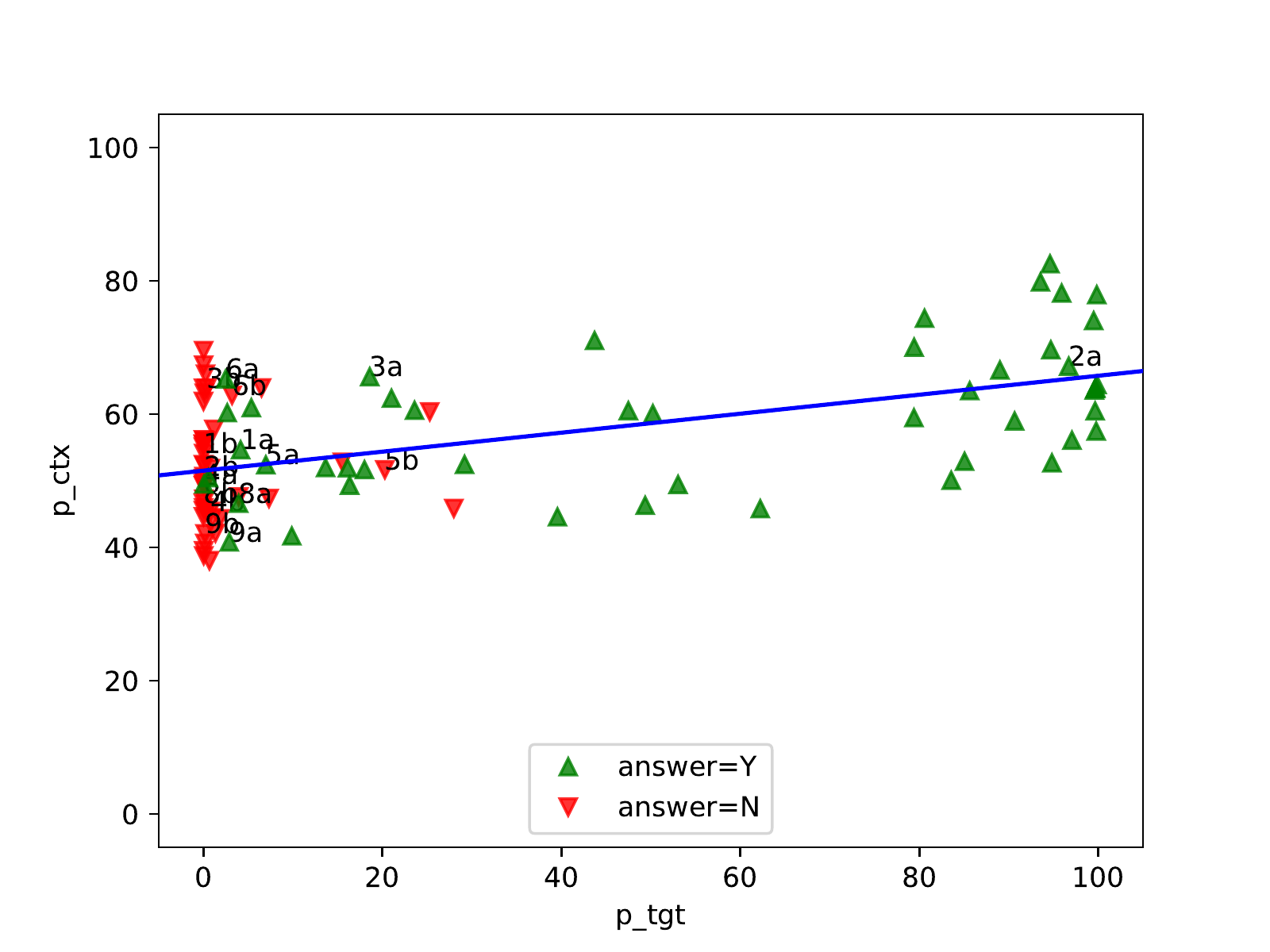}%
    \includegraphics[width=0.33\textwidth]{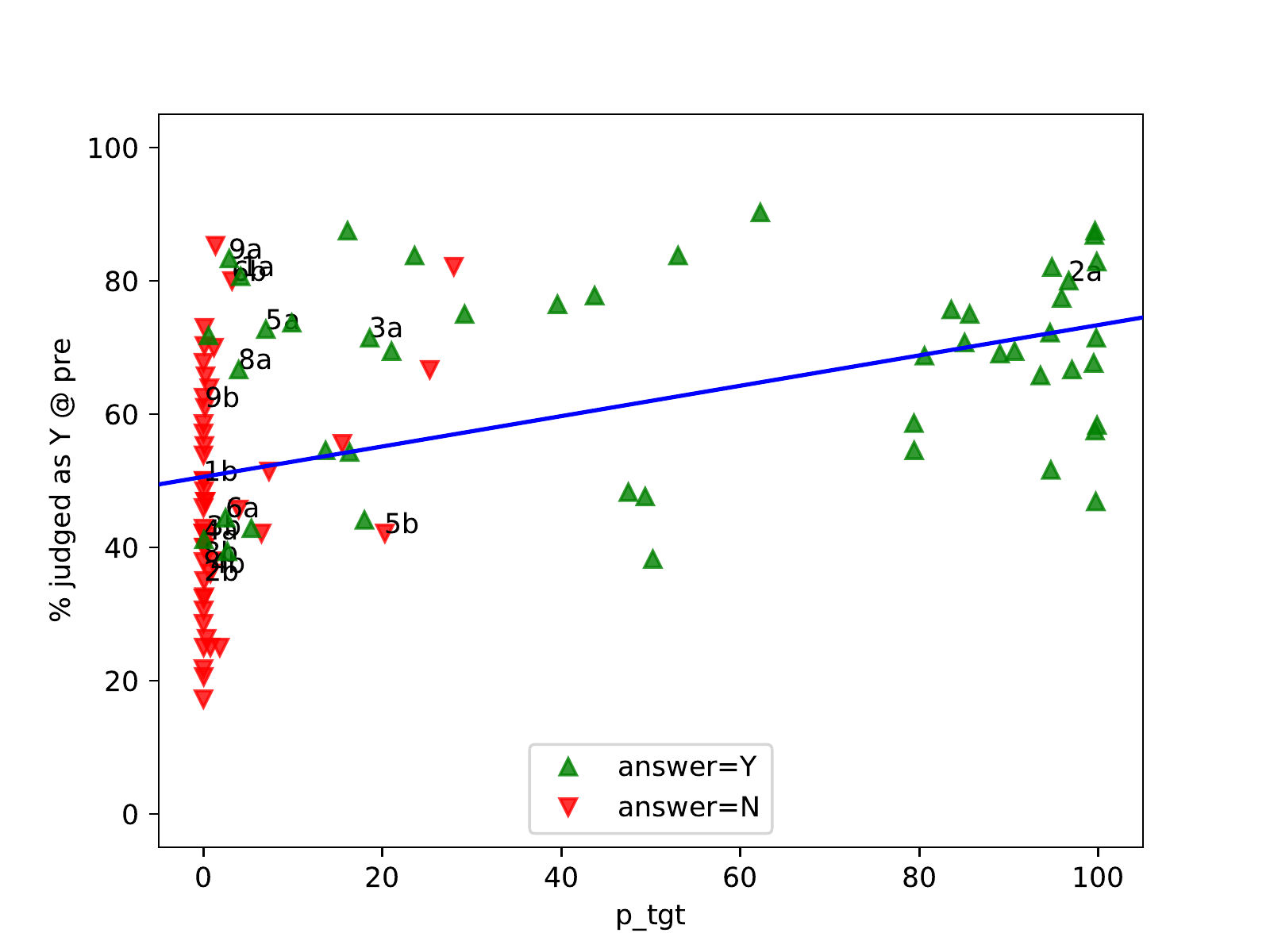}%
    \includegraphics[width=0.33\textwidth]{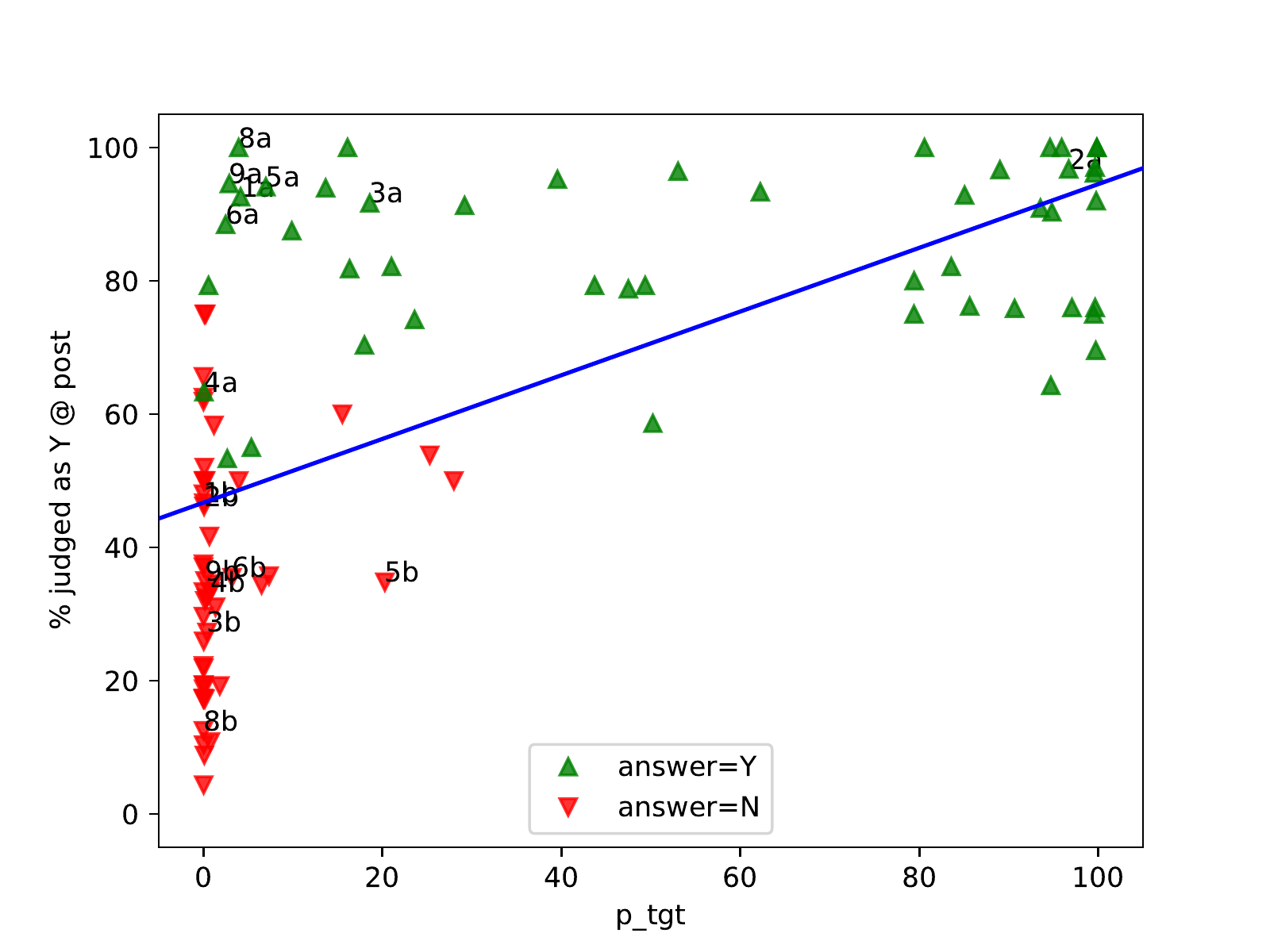}
    \caption{Correlations of LM probabilities, student grammaticality judgment \%, and intended answer (color\slash shape) for individual stimuli (points). \textbf{Left:}  $p_{tgt}$ ($x$) and $p_{ctx}$ ($y$);
    \textbf{Center:} $p_{tgt}$ ($x$) and pretest judgment ($y$); 
    \textbf{Right:} $p_{tgt}$ ($x$) and posttest judgment of non-control groups only ($y$). 
    $R^2$(answer, judge@post)=0.72; $R^2$($p_{tgt}$, answer)=0.48; $R^2$($p_{tgt}$, judge@post, blue line right)=0.41; $R^2$($p_{tgt}$, $p_{ctx}$, blue line left)=0.30; $R^2$(answer, judge@pre)=0.28; $R^2$($p_{tgt}$, judge@pre, blue line center)=0.22; $R^2$($p_{ctx}$, answer)=0.15; $R^2$($p_{ctx}$, judge@post)=0.13; $R^2$($p_{ctx}$, judge@pre)=0.04.
    Data points\slash sentences listed in \cref{tab:examples-probs} and discussed in \cref{sec:exp-lm-prob} are labeled.}
    \label{fig:lm-results}
\end{figure*}


\noindent\textbf{LM probabilities.}
The model was trained on GJT data only.
Recall from \cref{sec:mod-factors} that GJT testing prompts did not explicate the target preposition or even mention the word `preposition'.
All else equal, it is thus conceivable that, despite the preposition-focused training, students evaluate the sentences' grammaticality for reasons unrelated to the target preposition.
However, we can with high probability rule out this option as our model estimates strong effects for numerous features directly related to the preposition, namely: 
$p_{tgt}$ by itself ($d$=4.57; $p$<0.0001***); interaction $p_{tgt}$:$p_{ctx}$ ($d$=14.92; $p$<0.0001***);\footnote{While $p_{ctx}$ by itself is only very weakly correlated with either grammaticality or student response, it does become a useful predictor in interaction with $p_{tgt}$ (cf. \cref{fig:lm-results} left).}
and spatial vs.\ abstract usage of each preposition form and function (\cref{fig:effect-results-repl}, \cref{fig:prep-usage} in \cref{app:detailed-results}).
%
%
%
%
%
Furthermore, due to the heavy interaction between LM probabilities and categorical cue properties,\footnote{Linguistic categories may to some extent be encoded in the LM's distributed representations \citep{jawahar-etal-2019-bert}.} the singular random effect of spatial vs.\ abstract usage decreases when the model considers the LM-based fixed effects ($d$=0.372; $p$=0.0093**) compared to when it does not ($d$=0.491; $p$=0.0006***, \cref{fig:effect-results-repl}).

\subsection{Predicting Student Responses}\label{sec:prediction}

\noindent\textbf{Setup.}
We train the BLM and MLP using a training:evaluation:development data split ratio of 84:15:1, placing less weight on the dev set since it is only used to determine early-stopping during MLP training.
Experiments are run 10 times with random data splits and model initializations. 

\begin{table}[t]
    \setlength{\tabcolsep}{3pt}
    \centering\small
    \begin{tabular}{l cc cc}
         & \multicolumn{2}{c}{GJT + PET} & \multicolumn{2}{c}{GJT only} \\\cmidrule(r){2-3}\cmidrule(lr){4-5}
         & \textbf{BLM} & \textbf{MLP} & \textbf{BLM} & \textbf{MLP} \\\midrule
Uniform BL & \multicolumn{2}{c}{49.7 \textcolor{gray}{\scriptsize $\pm$1.1}} & \multicolumn{2}{c}{49.7 \textcolor{gray}{\scriptsize $\pm$1.2}} \\ 
BLM prior BL & \multicolumn{2}{c}{49.7 \textcolor{gray}{\scriptsize $\pm$2.1}} & \multicolumn{2}{c}{48.2 \textcolor{gray}{\scriptsize $\pm$1.4}} \\ 
Majority BL & \multicolumn{2}{c}{64.2 \textcolor{gray}{\scriptsize $\pm$0.9}} & \multicolumn{2}{c}{68.1 \textcolor{gray}{\scriptsize $\pm$0.7}} \\\midrule 
Full model & \underline{72.6} \textcolor{gray}{\scriptsize $\pm$1.1} & 71.5 \textcolor{gray}{\scriptsize $\pm$0.6} & \underline{72.5} \textcolor{gray}{\scriptsize $\pm$0.8} & 71.3 \textcolor{gray}{\scriptsize $\pm$0.9} \\ 
$-$ students & $-$2.2 \textcolor{gray}{\scriptsize $\pm$0.6} & $-$0.9 \textcolor{gray}{\scriptsize $\pm$0.7} & \textbf{$-$2.6} \textcolor{gray}{\scriptsize $\pm$0.9} & \textbf{$-$2.0} \textcolor{gray}{\scriptsize $\pm$0.8} \\ 
$-$ answer & \textbf{$-$5.6} \textcolor{gray}{\scriptsize $\pm$0.8} & \textbf{$-$4.6} \textcolor{gray}{\scriptsize $\pm$0.6} & $-$2.4 \textcolor{gray}{\scriptsize $\pm$0.8} & \textbf{$-$2.0} \textcolor{gray}{\scriptsize $\pm$0.8} \\ 
$-$ fxn \& usage & $-$5.4 \textcolor{gray}{\scriptsize $\pm$1.0} & \textbf{$-$4.6} \textcolor{gray}{\scriptsize $\pm$1.0} & $-$1.5 \textcolor{gray}{\scriptsize $\pm$0.4} & $-$0.8 \textcolor{gray}{\scriptsize $\pm$1.3} \\ 
$-$ instr \& time & $-$2.1 \textcolor{gray}{\scriptsize $\pm$0.9} & $-$1.8 \textcolor{gray}{\scriptsize $\pm$0.9} & $-$0.4 \textcolor{gray}{\scriptsize $\pm$0.7} & $-$1.4 \textcolor{gray}{\scriptsize $\pm$0.9} \\ 
$-p_{tgt}$ \& $p_{ctx}$ & n/a & n/a & $-$0.9 \textcolor{gray}{\scriptsize $\pm$0.9} & $-$0.4 \textcolor{gray}{\scriptsize $\pm$0.9} \\ 
    \bottomrule
    \end{tabular}
    \caption{Baselines (BL), BLM and MLP prediction performance, and feature ablation (student response correctness prediction accuracy in \%). Means and standard deviations over 10 random seeds, which affect not only model initialization but also data splitting and shuffling. Best full model results on each data split are underlined; highest-impact features in each column are bolded.}
    \label{tab:results-blm-mlp-abl}
\end{table}

\noindent\textbf{Results.}
As shown in \cref{tab:results-blm-mlp-abl}, both models easily outperform simple baselines, and the two models' overall accuracies are roughly on par (within each other's stdevs) with a slight advantage for the BLM.
For predicting GJT outcomes only, the aforementioned interaction between students and answer types is most crucial, followed by information about the target preposition (BLM) and instruction (MLP), respectively.
The LM-based features $p_{tgt}$ and $p_{ctx}$ are useful for both models, but less so than the categorical ones. This is somewhat unexpected based on their strong effect sizes (\cref{sec:replication}) and the overwhelmingly high performance of LMs on other tasks. 
A potential reason is the contrast between the LM reflecting a gross average of language use---which indeed correlates with grammaticality ($R^2=0.48$, \cref{fig:lm-results})---and the unreliability of student judgments, especially at the pretest and in the control group (\cref{fig:effect-results-repl} top).
%
The lack of stimulus sentences (and thus LM probabilities) in the PET further increases the importance of the answer, form-function, and usage features in the GJT+PET condition.
We also see a larger ablation effect of the instruction and time features, which is consistent with the larger interaction effect estimates for the PET (\cref{fig:effect-results-repl} bottom).

\subsection{Qualitative Analysis of Stimuli}\label{sec:lm-prob}\label{sec:exp-lm-prob}

We take a closer look at individual stimuli in \cref{fig:lm-results}.
%
%
From the $y$-axis distribution in the center and right panels we can clearly see the learning development among students undergoing preposition-focused training.
At the pretest (center), aggregate students' grammaticality judgment is less decisive (mostly vertically centered around 50\%$\pm \approx 20pp$.
At the posttest (right), the spread is much more decisive, ranging from almost 0\% to 100\%.
At both points in time, there is a slight bias towards positive judgment, i.e., students are generally more willing to accept ungrammatical stimuli as grammatical than to reject grammatical ones. 
In contrast, LM probabilities ($x$-axis) tend to err on the conservative side, i.e., the LM has higher recall on recognizing ungrammatical items, whereas students have higher recall on recognizing grammatical items, each at the cost of precision.\footnote{Note that LM probabilities are not based on a binary grammaticality decision but on a selection decision over the entire vocabulary, and also that gradient linguistic judgments in general cannot be said to \textit{only} revolve around grammaticality \citep[cf.][]{lau2017grammaticality}. We could address this by looking at the ratio between the probabilities for each pair, but that would in turn establish a dependency among stimuli within each pair which is not present in the human experiment---each stimulus is presented in isolation, in randomized order. Thus, for transparency, we stick with the plain probability and elaborate qualitatively on the expected behavior below.}

We expect that intended-grammatical ($\checkmark$) usages generally receive higher LM probabilities ($\Delta p$) than intended-ungrammatical (\ding{55}) usages.
This is the case most of the time (for 41/48 stimulus pairs total), except for 7 cases, 6 of which involve the preposition `\textit{over}' as the target.
We present these sentences in \cref{tab:examples-probs}, along with 3 examples where both $\Delta p$'s are as expected.

What makes ex. 4 -- 9 special?
A potential explanation is that the verb+preposition+object constructions in ex. 1 -- 3 seem to be more clearly distinguishable as either grammatical or ungrammatical than the rest.
In contrast, the \ding{55} sentences in ex. 4 -- 6 are not \textit{truly} ungrammatical. The scenarios they describe are unlikely but possible, and the unlikeliness mostly arises through the full-sentence context rather than the prepositional construction alone. In fact, each alternative preposition in 4b, 5b, and 6b might in isolation be a \textit{more} expected collocation with the verb than `\textit{over}', which would explain the $p_{tgt}$ trend.
Ex. 7 -- 9 (both \ding{55} and $\checkmark$) describe much more rare (i.e., unlikely as far as the distributional LM is concerned) scenes, which may lead to the overall lower $p_{ctx}$ values.\footnote{A second tendency may lie in the concreteness and perceived simplicity (both in terms of semantics and register) of the preposition-governing \textit{verbs}: `\textit{hang, watch, fall}' are all fairly concrete, unambiguous, and colloquial, whereas `\textit{reach, diffuse, stretch, sweep}' have more specialized meanings and are somewhat higher register.}

%
%
%
%

%



\section{Discussion}\label{sec:discussion}


\subsection{Which model type is most appropriate?}\label{sec:model-discussion}
For the purpose of our study, the Bayesian logistic model of student responses has clear advantages over both the previous frequentist analysis of score aggregates (complexity of interactions, intuitiveness; \cref{sec:replication}) and the neural response classifier (higher interpretability with roughly equal prediction accuracy; \cref{sec:prediction}).
However, while this observation is in line with both our expectations and recent literature in SLA \citep[e.g.,][]{norouzian2018bayesian,norouzian2019bayesian}, we still recommend testing model practicability on a case-by-case basis.
For example, if much more training data is available, a neural classifier is likely to outperform a sparse model at prediction accuracy.
Whenever the BLM and ANOVA agree on a feature's significance (and they usually---but not always---do), the BLM's estimates are relatively amplified (\cref{sec:replication}). This can be useful for \textit{identifying} potentially relevant effects and interactions, but should also be taken with a grain of salt as it sometimes may construe results too optimistically.
Where do these divergences come from? We hesitate to make any strong statements about broad philosophical differences between Bayesian and frequentist statistics in the abstract.
Rather, we suspect that it mostly comes down to practical considerations like framing model and data around individual item responses vs.\ aggregate score, as well as varying degrees of commitment to latent sampling and optimization.
Item response prediction accuracy and ablation analyses give some insight into how individual features affect models' estimates of the outcome variable and is consistent with statistical analyses (\cref{sec:prediction}).
This is particularly useful for discriminative neural models such as our MLP classifier, and is, of course, common practice in NLP classification studies.
However, it is also much more costly, less precise, and less reliable than Bayesian and frequentist approaches.

\subsection{Implications for SLA}\label{sec:sla-discussion}
Our analysis of answer types and student aptitudes (\cref{sec:replication,sec:prediction}) confirms \citeposs{wong2022fostering} and others' findings about differences between productive and receptive knowledge. We support \citeauthor{wong2022fostering}'s argument that the type of assessment should align with both instruction type and and intended learning outcome.
We further observe that even within the generally \textit{receptive} task of grammaticality judgment, the subtask of ruling out ungrammatical items (GJT\mbox{-}N) requires higher specificity than accepting grammatical ones (GJT\mbox{-}Y) and is thus more closely aligned with \textit{productive} tasks (e.g., PET).
Interestingly, students who are better than average at productive tests tend to be slightly weaker than average at receptive ones and vice versa.
A potential future use case of explicitly modeling students' individual differences w.r.t.\ different task types and linguistic items is that educational applications can be tailored to their \textit{weaknesses}, which is expected to increase learning effectiveness and efficiency.\footnote{In practice, such a process should ideally be decentralized by training separate models for each student on the client side, to uphold privacy and other ethical standards.}
Outside of directly deploying learning technology to end users, our findings can inform educators and SLA researchers.
For example, unexpected patterns in LM probabilities (\cref{sec:lm-prob}) may point to suboptimally designed stimulus pairs. Thus, LM probing could be a useful tool in cue selection and stimulus design of similar studies in the future.

\subsection{Implications for NLP}\label{sec:nlp-discussion}
In this work, we primarily analyze human learner behavior \textit{using} different machine learning models, while in NLP-at-large it is much more common to \textit{analyze} machine learning models w.r.t.\ a human ground truth.
At the same time, our observations that different senses and usages even of the same preposition form heavily affect human learnability are somewhat analogous to previous results in automatic preposition disambiguation \citep[varying model performance for extended vs.\ lexicalized senses;][]{schneider2018comprehensive,liu-etal-2019-linguistic}.
\citeauthor{liu-etal-2019-linguistic} also found that LM pretraining improves disambiguation performance, while \citet{kim-etal-2019-probing} drew attention to differences among various NLP tasks as `instruction methods'.
This is not to say that current LM training practices are necessarily plausible models of human language learning and teaching, but even these high-level similarities in behavioral patterns invite further investigation.

\section{Conclusion}\label{sec:conclusions}

Much quantitative research in many areas of linguistics, including SLA, has been relying on the frequentist method for a long time---and for good reasons: It enables strong conclusions about clear hypotheses, closely following the observed data.

Here we compared several alternative approaches to estimating a multitude of potential effects more holistically, namely via IRT-inspired Bayesian sparse models of explicit interactions among facts, neural classifiers of student responses and feature ablation, as well as contextual probabilities of the experimental stimuli obtained from a pretrained language model (\cref{sec:models}).

Overall, we were able to replicate previous frequentist findings regarding the difficulty of acquiring the preposition system in English as a second language and the benefits of concept-based instruction (\cref{sec:replication}).
Our computational analysis emphasized the increased flexibility and occasionally stronger effect size estimates of IRT and Bayesian models, as well as their natural interpretability compared to neural models with equal predictive power.


We also found novel interactions among task and subtask type, student individual differences, preposition cue and LM contextualization (\cref{sec:exp}), and discussed them in the broader contexts of both NLP and SLA, hoping to build bridges between the two research communities (\cref{sec:discussion}).
As a final take-away for both fields, the differences between the LM's and students' overall tendencies to accept or reject stimuli (\cref{sec:exp-lm-prob,fig:lm-results} right) could potentially be exploited in both directions: The aggregate distributional grammatical knowledge of an LM could be used to teach students the most accepted usages of prepositions and other function words across a large population of speakers (i.e., improve their specificity), while LMs could learn to be more creative and to utilize humans' intuitive cross-lingual meaning mappings by learning from second-language learner data.

\section*{Limitations}

Our study and findings are limited to the specific L1--L2 pair of Chinese (Mandarin and Cantonese)--English.
Further, the experimental setting we draw our data from is highly controlled, with carefully-chosen lexical items and carefully-designed (length- and distractor-matched) stimulus sentences. 
While this enables strong statistical conclusions about the data itself, it poses a sparsity problem for most state-of-the-art NLP models, as can be seen even in the small and simple multi-layer perceptron we test.

While it would also be interesting to know whether students respond differently to the same instruction type or vice versa, the between-subjects experimental design underlying our data does not allow such a measurement.

We inspect several model types representing a selection of extreme areas of a vast continuum of computational analysis methodologies.
Naturally, this means that we cannot go into a lot of depth regarding model engineering and detailed comparison among similar implementations of each type.

\section*{Ethics Statement}

Student identities are completely anonymized in our analyses and in the data we feed to our models.
By locally distinguishing individual students, we do not wish to single out, over-interpret, or judge any individual student's behavior or aptitude, but rather to fit the models to our data as best we can and also to control for spurious patterns that might have been missed during initial outlier-filtering.

\section*{Acknowledgments}

We thank the anonymous reviewers for their insightful questions and feedback.
This work has been supported by Hong Kong PolyU grant 1\mbox{-}YWBW, awarded to the first author, and grant EDB(LE)/P\&R/EL/203 of the Hong Kong Standing Committee on Language Education and Research (SCOLAR), awarded to the second author.

\bibliography{main}
\bibliographystyle{acl_natbib}

\newpage

\appendix

\section{Effects of Preposition Cues}\label{app:detailed-results}

In the main text, for brevity, we omitted a detailed analysis of the effects of specific combinations of preposition form, function, and usage on student performance.
Here we take a closer look at the six types of cues: \textit{in} with the \textsc{Containment} function, \textit{at} with the \textsc{Target} and \textsc{Point} functions, and \textit{over} with the \textsc{Higher},\textsc{Cover}, and \textsc{Cross} functions.

\begin{figure}[h]
    \centering\small
    \includegraphics[width=0.87\columnwidth]{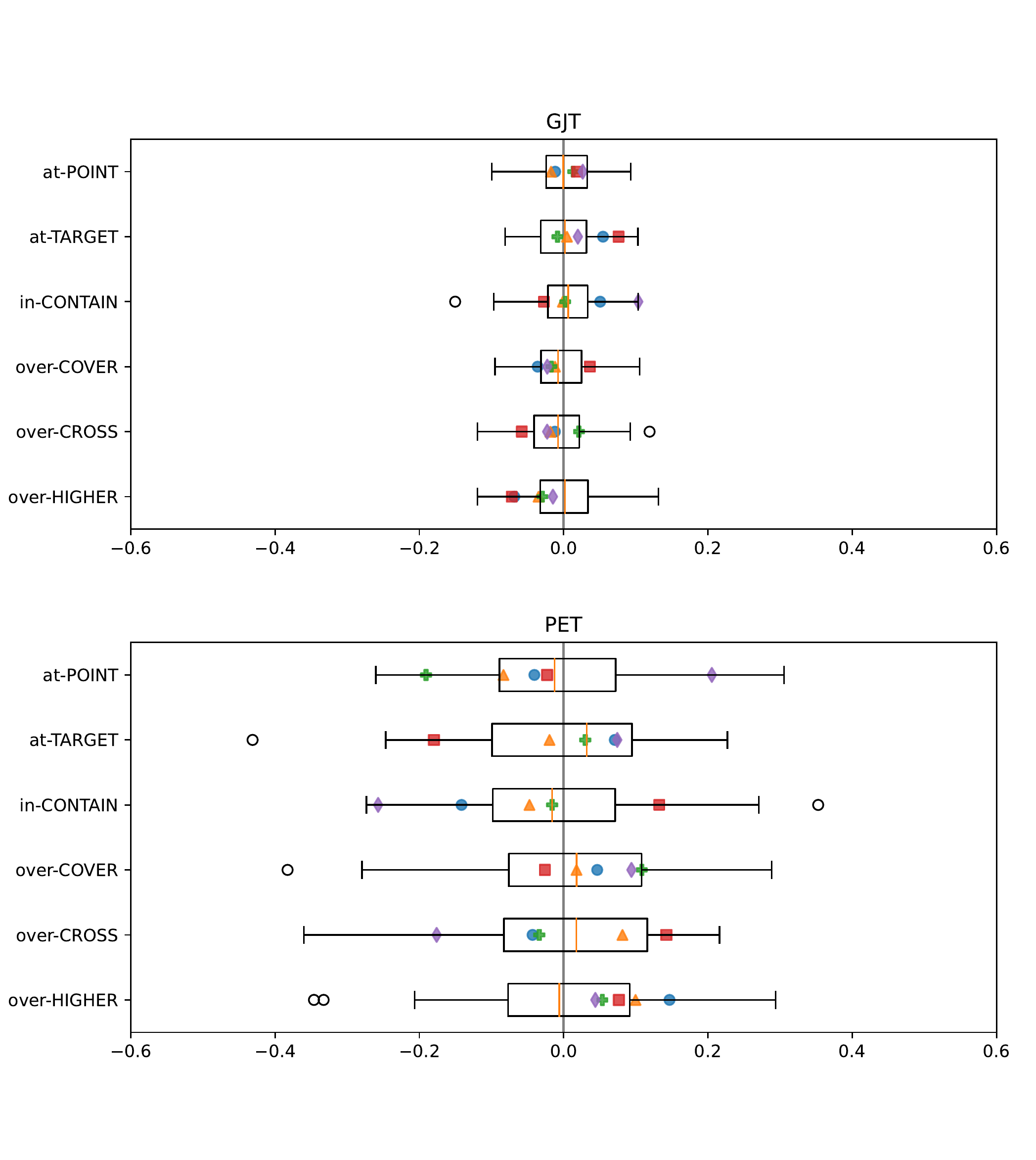}
    \caption{Spread among student effect means ($x$-axis) in interaction with preposition form\slash function. 5 randomly chosen students are shown exemplarily (filled shapes; empty circles are outliers). Note that, while in our other figures the error bars denote standard deviations over models' marginal parameter distributions, here they describe the distribution over students of estimated mean interaction effects.}
    \label{fig:stud-prep}
\end{figure}

In \cref{fig:stud-prep}, we see that there is a wide spread among students for each of the cue types, especially at the PET. The fact that these effects are estimated as interactions in addition to the student-level intercepts suggests, again, that students' skill sets are unique, depending on the preposition cue, which is also illustrated for 5 randomly chosen students.

\begin{sidewaysfigure*}
    \centering
    \includegraphics[width=\textwidth]{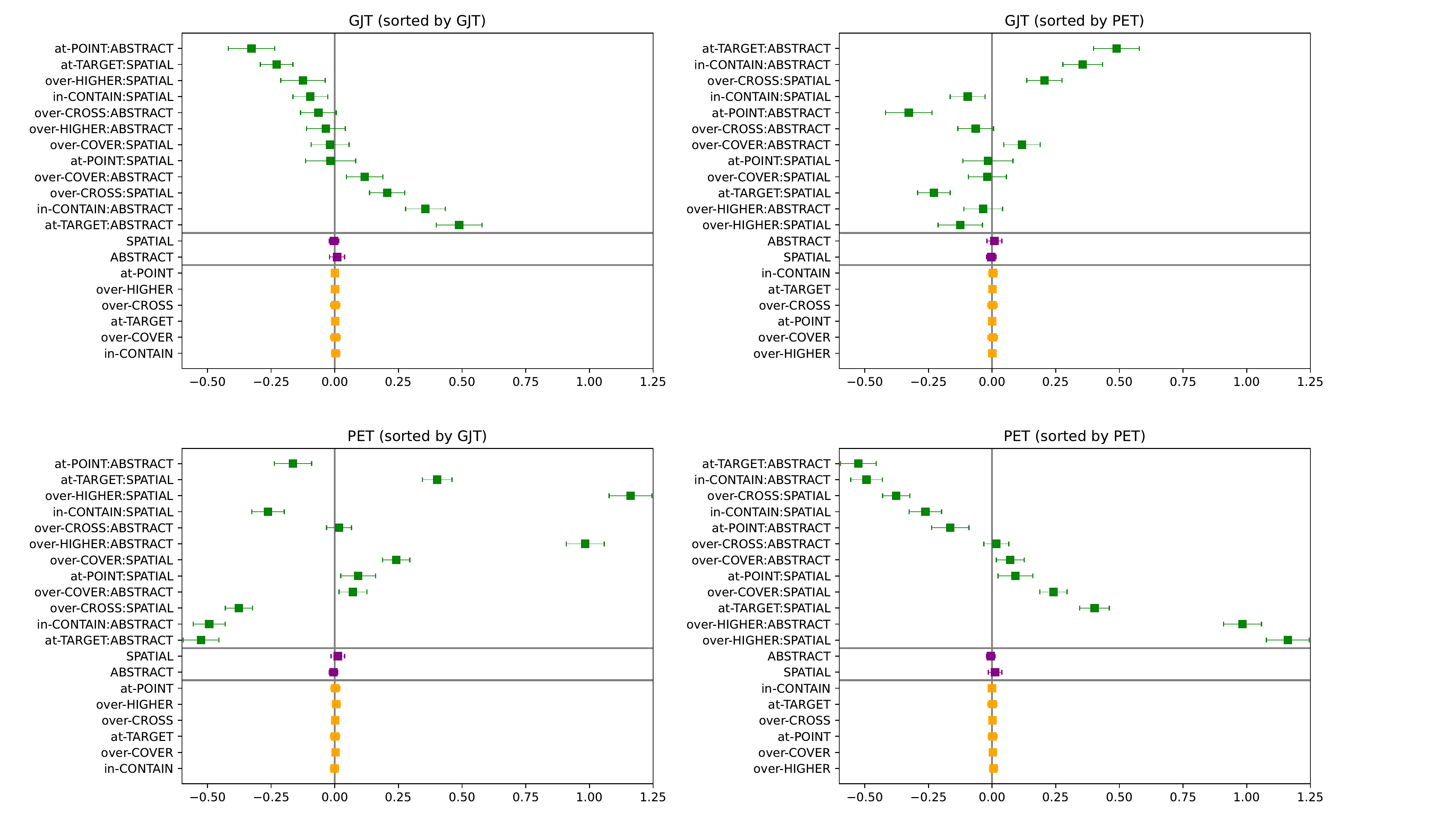}
    \caption{Effect estimates for interactions between preposition form\slash fxns and spatial vs.\ abstract usage.}
    \label{fig:prep-usage}
\end{sidewaysfigure*}

In \cref{fig:prep-usage}, we see that the difficulty of these six cues varies greatly, depending on both spatial\slash abstract use and task type. In fact, the difficulty ranking is largely reversed between GJT and PET.
As a striking example of this, \textit{at}-\textsc{Target}-Abstract and \textit{in}-\textsc{Contain}-Abstract are the easiest cues to judge correctly in the GJT but most difficult to produce in the PET.
There exceptions to this trend, too. E.g., \textit{at}-\textsc{Point}-Abstract is relatively difficult in both GJT and PET.
Another interesting observation is that, in the PET, both usages of \textit{over}-\textsc{Higher} are much easier to produce than any other cue.

\end{document}